\documentclass{tlp}

\usepackage{tkz-graph}
\usepackage{pgfplots}
\usepackage{bbm} % for indicator function
\usepackage{amssymb}
\usepackage{hyperref}
\usepackage{algorithm}
\usepackage{algpseudocode}

\usepackage{multirow}

\usepackage{subcaption}

\newcommand{\revisione}[1]{\textcolor{black}{#1}}

\definecolor{red_plot}{HTML}{F30522}
\definecolor{first_plot}{HTML}{F30522}

\definecolor{orange_plot}{HTML}{FA5F22}

\definecolor{yellow_plot}{HTML}{DED712}
\definecolor{fourth_plot}{HTML}{DED712}

\definecolor{pink_plot}{HTML}{FFD0B4}
\definecolor{light_blue_plot}{HTML}{D1E5F0}
\definecolor{blue_plot}{HTML}{67A9CF}

\definecolor{dark_blue_plot}{HTML}{2166AC}
\definecolor{third_plot}{HTML}{2166AC}

\definecolor{green_plot}{HTML}{20D525}
\definecolor{second_plot}{HTML}{20D525}

\newcommand{\resizeGraphFactor}{0.85}

\newcommand{\impl}{\ {:\!-}\  }

\DeclareMathOperator{\program}{\mathcal{P}}
\DeclareMathOperator{\interpretations}{\mathcal{I}}
\DeclareMathOperator{\lowerprob}{\underline{P}}
\DeclareMathOperator{\upperprob}{\overline{P}}
\DeclareMathOperator{\argmax}{arg\,max}

\DeclareMathOperator{\upperexp}{\overline{E}}

\newtheorem{example}{Example}
\newtheorem{definition}{Definition}

\usepackage{listings}
\begin{document}

\lefttitle{Cambridge Author}

\jnlPage{1}{14}
\jnlDoiYr{2024}
\doival{10.1017/xxxxx}

\title[Symbolic Parameter Learning in Probabilistic Answer Set Programming]{Symbolic Parameter Learning in Probabilistic Answer Set Programming}

\begin{authgrp}
\author{\gn{Damiano} \sn{Azzolini} }
\affiliation{Department of Environmental and Prevention Sciences -- University of Ferrara, Ferrara, Italy \\
\email{damiano.azzolini@unife.it}}
\author{\gn{Elisabetta} \sn{Gentili} }
\affiliation{Department of Engineering -- University of Ferrara, Ferrara, Italy \\
\email{e.gentili1@unife.it}}
\author{\gn{Fabrizio} \sn{Riguzzi} }
\affiliation{Department of Mathematics and Computer Science -- University of Ferrara, Ferrara, Italy\\
\email{fabrizio.riguzzi@unife.it}}
\end{authgrp}

\history{\sub{xx xx xxxx;} \rev{xx xx xxxx;} \acc{xx xx xxxx}}

\maketitle

\begin{abstract}
Parameter learning is a crucial task in the field of Statistical Relational Artificial Intelligence: given a probabilistic logic program and a set of observations in the form of interpretations, the goal is to learn the probabilities of the facts in the program such that the probabilities of the interpretations are maximized.
In this paper, we propose two algorithms to solve such a task within the formalism of Probabilistic Answer Set Programming, both based on the extraction of symbolic equations representing the probabilities of the interpretations.
The first solves the task using an off-the-shelf constrained optimization solver while the second is based on an implementation of the Expectation Maximization algorithm.
Empirical results show that our proposals often outperform existing approaches based on projected answer set enumeration in terms of quality of the solution and in terms of execution time.
The paper has been accepted at the ICLP2024 conference and is under consideration in Theory
and Practice of Logic Programming (TPLP).
\end{abstract}


\maketitle

\section{Introduction}
\label{sec:introduction}
Statistical Relational Artificial Intelligence (StarAI)~\citep{raedt2016statistical} is a subfield of Artificial Intelligence aiming at describing complex probabilistic domains with interpretable languages.
Such languages are, for example, Markov Logic Networks~\citep{richardson2006markov}, Probabilistic Logic Programs~\citep{Rig23-BKaddress,DBLP:conf/ijcai/RaedtKT07}, and Probabilistic Answer Set Programs~\citep{cozman2020pasp}.
Here, we focus on the last.
Within StarAI, there are many problems that can be considered such as probabilistic inference, MAP inference, abduction, parameter learning, and structure learning. 
In particular, the task of parameter learning requires, given a probabilistic logic program and a set of observations (often called interpretations), tuning the probabilities of the probabilistic facts such that the likelihood of the observation is maximized.
This is often solved by means of Expectation Maximization (EM)~\citep{BelRig13-IDA-IJ,AzzBelRig2022ParameterLearningPASP,dries2015problog2} or Gradient Descent~\citep{DBLP:conf/pkdd/GutmannKKR08}.

Recently,~\cite{AzzRig21-ICLP-IC} proposed to extract a symbolic equation for the probability of a query posed to a probabilistic logic program.
From that equation, it is possible to model and solve complex constrained problems involving the probabilities of the facts~\citep{azz2023ILP}.

In this paper, we propose two algorithms to learn the parameters of probabilistic answer set programs.
Both algorithms are based on the extraction of symbolic equations from a compact representation of the interpretations, but they differ in how they solve the problem.
The first casts parameter learning as a nonlinear constrained optimization problem and leverages off-the-shelf solvers, thus bridging the area of probabilistic answer set programming with constrained optimization, while the second solves the problem using EM.
Empirical results, also against an existing tool to solve the same task, on four different datasets with multiple configurations show the proposal based on constrained optimization is often significantly faster and more accurate w.r.t. the other approaches.

The paper is structured as follows: Section~\ref{sec:background} discusses background knowledge, Section~\ref{sec:algorithm} proposes novel algorithms to solve the parameter learning task, that are tested in Section~\ref{sec:experiments}.
Section~\ref{sec:related} surveys related works and Section~\ref{sec:conclusions} concludes the paper.

\section{Background}
\label{sec:background}
\revisione{An answer set program is composed by a set of normal rules of the form 
$h \impl b_0, \dots, b_m, \ not \ c_0, \dots, \ not\ c_n$
where $h$, the $b_i$s, and the $c_i$s are \textit{atoms}.
$h$ is called \textit{head} while the conjunction of literals after the ``$\impl$'' symbol is called \textit{body}.
A rule without a head is called \textit{constraint} while a rule without a body is called \textit{fact}.
The semantics of an answer set program is based on the concept of \textit{stable model}~\citep{gelfond1988stable}, often called \textit{answer set}.
The set of all possible ground atoms for a program $P$ is called \textit{Herbrand base} and denoted with $B_P$.
The grounding of a program $P$ is obtained by replacing variables with constants in $B_P$ in all possible ways.
An \textit{interpretation} is a subset of atoms of $B_P$ and it is called \textit{model} if it satisfies all the groundings of $P$.
An \textit{answer set} $I$ of $P$ is a minimal model under set inclusion of the reduct of $P$ w.r.t. $I$, where the reduct w.r.t. $I$ is obtained by removing from $P$ the rules whose body is false in $I$.
% We do not add further details about Logic Programming and Answer Set programming and refer the reader to~\citep{Lloyd87,brewka2011asp}. 
}

\subsection{Probabilistic Answer Set Programming}
\label{subsec:pasp}
We consider the Credal Semantics (CS)~\citep{DBLP:conf/ilp/CozmanM16,cozman2020complexity,DBLP:conf/ecsqaru/Lukasiewicz05} that associates a meaning to Answer Set Programs extended with probabilistic facts~\citep{DBLP:conf/ijcai/RaedtKT07} of the form $p::a$ where $p$ is the probability associated with the atom $a$.
Intuitively, such notation means that the fact $a$ is present in the program with probability $p$ and absent with probability $1-p$.
These programs are called Probabilistic Answer Set Programs (PASP, and we use the same acronym to also indicate Probabilistic Answer Set Programming -- the meaning will be clear from the context).
A selection for the presence or absence of each probabilistic fact defines a world $w$ whose probability is $P(w) = \prod_{a \in w} p \cdot \prod_{a \not \in w} (1-p)$,
where with $a \in w$ we indicate that $a$ is present in $w$ and with $a \not \in w$ that $a$ is absent in $w$.
\revisione{A program with $n$ probabilistic facts has $2^n$ worlds.
Let us indicate with $W$ the set of all possible worlds.}
Each world is an answer set program and it may have zero or more answer sets but the CS requires at least one.
If this holds, the probability of a query $q$ (a conjunction of ground literals) is defined by a lower and an upper bound.
A world $w$ contributes to both the lower and upper probability if \revisione{each of its answer sets contains the query, i.e., it is a cautious consequence}.
If \revisione{only some answer sets contain the query, i.e., it is a brave consequence,} $w$ only contributes to the upper probability.
If the query is not present, we have no contribution to the probability bounds from $w$. 
In formulas,
\begin{equation}
\label{eq:upper_lower_prob}
P(q) 
  = [\lowerprob(q),\upperprob(q)]
  = [\sum_{w_i \in W \mid \forall m \in AS(w_i), \ m \models q} P(w_i), \sum_{w_i \in W \mid \exists m \in AS(w_i), \ m \models q} P(w_i)].
\end{equation}
The conditional probability of a query $q$ given evidence $e$, also in the form of conjunction of ground literals, is~\citep{cozman2020pasp}:
\begin{equation}
\label{eq:conditional_probability}
\lowerprob(q \mid e) = \frac{\lowerprob(q, e)}{\lowerprob(q, e) + \upperprob(not \ q, e)}, \ 
\upperprob(q \mid e) = \frac{\upperprob(q, e)}{\upperprob(q, e) + \lowerprob(not \ q, e)}. 
\end{equation}
For the lower conditional probability $\lowerprob(q \mid e)$, 
if $\lowerprob(q, e) + \upperprob(not \ q, e) = 0$ and $\upperprob(q, e) > 0$, \revisione{then} $\lowerprob(q \mid e) = 1$.
Similarly, for the upper conditional probability $\upperprob(q \mid e)$,
if $\upperprob(q, e) + \lowerprob(not \ q, e) = 0$ and $\upperprob(not \ q, e) > 0$, \revisione{then} $\upperprob(q \mid e) = 0$.
Both formulas are undefined if $\upperprob(q, e)$ and $\upperprob(not \ q, e)$ are 0.

To clarify, consider the following example.
\begin{example}
\label{ex:running_pasp}
The following PASP encodes a simple graph reachability problem.
\begin{lstlisting}
0.2::edge(1,2).
0.3::edge(2,4).
0.9::edge(1,3).
path(X,Y):- connected(X,Z), path(Z,Y).
path(X,Y):- connected(X,Y).

connected(X,Y):- edge(X,Y), not nconnected(X,Y).
nconnected(X,Y):- edge(X,Y), not connected(X,Y).
\end{lstlisting}
The first three facts are probabilistic.
The rules state that there is a path between $X$ and $Y$ if they are directly connected or if there is a path between $Z$ and $Y$ and $X$ and $Z$ are connected.
Two nodes may or may not be connected if there is an edge between them.
There are $2^3 = 8$ worlds to consider, listed in Table~\ref{tab:world_running}.
If we want to compute the probability of the query $q = path(1,4)$, only $w_6$ and $w_7$ contribute to the upper bound (no contribution to the lower bound), obtaining $P(q) = [0,0.06]$.
If we observe $e = edge(2,4)$, we get $P(q,e) = [0,0.06]$, $P(not \ q,e) = [0.24,0.3]$, thus $\lowerprob(q \mid e) = 0$ and $\upperprob(q \mid e) = 0.2$.
\end{example}

\begin{table}
\caption{Worlds and their probabilities for Example~\ref{ex:running_pasp}.
The second, third, and fourth columns contain 0 or 1 if the corresponding probabilistic fact is respectively false or true in the considered world.
The LP/UP column indicates whether the considered world \revisione{contributes to the lower (LP) or upper (UP) bound} (or does not contribute, marked with a dash) for the probability of the query $path(1,4)$.}
\label{tab:world_running}
\centering
 {\tablefont\begin{tabular}{@{\extracolsep{\fill}}cccccc}
   \topline
%\begin{tabular}{| c | c | c | c | c | c |}
  $w_{id}$ & $edge(1,2)$ & $edge(2,4)$ & $edge(1,3)$ & LP/UP & Probability\midline
 % \hline
  $w_{0}$ & 0 & 0 & 0 & -  & 0.056 \\
  $w_{1}$ & 0 & 0 & 1 & -  & 0.504 \\
  $w_{2}$ & 0 & 1 & 0 & -  & 0.024 \\
  $w_{3}$ & 0 & 1 & 1 & -  & 0.216 \\
  $w_{4}$ & 1 & 0 & 0 & -  & 0.014 \\
  $w_{5}$ & 1 & 0 & 1 & -  & 0.126 \\
  $w_{6}$ & 1 & 1 & 0 & UP & 0.006 \\
  $w_{7}$ & 1 & 1 & 1 & UP & 0.054 
\botline 
\end{tabular}}
\end{table}

Inference in PASP can be expressed as a Second Level Algebraic Model Counting Problem (2AMC)~\citep{DBLP:journals/tplp/KieselTK22}.
Given 
a propositional theory $T$ where its variables are grouped into two disjoint sets, $X_o$ and $X_i$,
two commutative semirings $R^{i} = (D^i, \oplus^i, \otimes^i, n_{\oplus^i}, n_{\otimes^i})$ and $R^{o} = (D^o, \oplus^o, \otimes^o, n_{\oplus^o}, n_{\otimes^o})$,
two weight functions, $w_i : lit(X_i) \to D^i$ and $w_o : lit(X_o) \to D^o$, 
and a transformation function $f : D^i \to D^o$, 
2AMC is encoded as:
\begin{equation*}
  % \label{eq:2amc}
  % \begin{split}  
    2AMC(T) =
    \bigoplus\nolimits_{I_{o} \in \mu(X_{o})}^{o} 
    \bigotimes\nolimits^{o}_{a \in I_{o}} 
    w_{o}(a) 
    \otimes^{o}
    f(
      \bigoplus\nolimits_{I_{i} \in \varphi(T \mid I_{o})}^{i} \bigotimes\nolimits^{i}_{b \in I_{i}} w_{i}(b)  
    )  
  % \end{split}
\end{equation*}
where $\varphi(T \mid I_{o})$ is the set of assignments to the variables in $X_i$ such that each assignment, together with $I_o$, satisfies $T$ and $\mu(X_{o})$ is the set of possible assignments to the variables in $X_{o}$.
\revisione{In other words, 2AMC requires to solve two Algebraic Model Counting (AMC)~\citep{10.1016/j.jal.2016.11.031} tasks. The outer task focuses on the variables $X_o$, and for each possible assignment to these variables, an inner AMC task is performed by considering $X_i$.
These two tasks are connected via a transformation function that turns values from the inner task into values for the outer task. }
Different instantiations of the components allow different tasks to be represented.
To perform inference in PASP,~\cite{AzzRig2023-AIXIA-IC} proposed to consider 
as innermost semiring $\mathcal{R}^{i} = (\mathbb{N}^2, +, \cdot, (0,0), (1,1))$ with $X_i$ containing the atoms of the Herbrand base except the probabilistic facts and $w_i$ mapping $not \ q$ to $(0, 1)$ and all other literals to $(1, 1)$,
as outer semiring the two-dimensional probability semiring, i.e., $\mathcal{R}^{o} = ([0, 1]^2, +, \cdot,(0, 0),(1, 1))$, with $X_o$ containing the atoms of the  probabilistic facts and $w_o$ associating $(p, p)$ and $(1 - p, 1 - p)$ to $a$ and $not \ a$, respectively, for every probabilistic fact $p :: a$ and $(1, 1)$ to all the remaining literals, and
as transformation function $f((n_1,n_2))$ returning the pair $(v_{lp},v_{up})$ where $v_{lp} = 1$ if $n_1 = n_2$, 0 otherwise, and $v_{up} = 1$ if $n_1 > 0$, 0 otherwise.

2AMC can be solved via knowledge compilation~\citep{DBLP:journals/jair/DarwicheM02}, often adopted in probabilistic logical settings, since it allows us to compactly represent the input theory with a tree or graph and then computing the probability of a query by traversing it.
aspmc~\citep{DBLP:conf/kr/EiterHK21,EITER2024104109} is one such tool, that has been proven more effective than other tools based on alternative techniques such as projected answer set enumeration~\citep{AzzBellRig2022PASTA,AzzRig2023-AIXIA-IC}.
aspmc converts the input theory into a negation normal form (NNF) formula, a tree where each internal node is labelled with either a conjunction (and-node) or a disjunction (or-node), and leaves are associated with the literals of the theory.
\revisione{More precisely, aspmc targets sd-DNNFs which are NNFs with three additional properties:
i) the variables of children of and-nodes are disjoint (decomposability property);
ii) the conjunction of any pair of children of or-nodes is logically inconsistent (determinism property); and
iii) children of an or-node consider the same variables (smoothness property).
Furthermore, aspmc also requires $X$-firstness~\citep{DBLP:journals/tplp/KieselTK22}, a property that constraints the order of appearance of variables: given two disjoint partitions $X$ and $Y$ of the variables in an NNF $n$, a node is termed pure if all variables departing from it are members of either $X$ or $Y$.
If this does not hold, the node is called mixed.
An NNF has the $X$-firstness property if for each and-node, all of its children are pure nodes or if one child is mixed and all the other nodes are pure with variables belonging to $X$.}

\subsection{Parameter Learning In Probabilistic Answer Set Programs}
\label{subsec:pl_pasp}
We adopt the same Learning from Interpretations framework of~\citep{AzzBelRig2022ParameterLearningPASP}, that we recall here for clarity.
We denote a PASP with $\program(\Pi)$, where $\Pi$ is the set of parameters that should be learnt.
The parameters are the probabilities associated to (a subset of) probabilistic facts.
We call such facts as \textit{learnable facts}.
Note that the probabilities of some probabilistic facts can be fixed, i.e., there can be some probabilistic facts that are not learnable facts.
A partial interpretation $I = \langle I^+,I^- \rangle$ is composed by two sets $I^+$ and $I^-$ that respectively represent the set of true and false atoms. 
It is called partial since it may specify the truth value of some atoms only.
Given a partial interpretation $I$, we call the interpretation query $q_I = \bigwedge_{i^+ \in I^+} i^+ \bigwedge_{i^- \in I^-} not \ i^-$.
The probability of an interpretation $I$, $P(I)$, is defined as the probability of its interpretation query, which is associated with a probability range since we interpret the program under the CS.
Given a PASP $\program(\Pi)$, the lower and upper probability for an interpretation $I$ are defined as
\begin{equation*}
% \label{eq:upper_prob_interpretation}
\upperprob(I \mid \program(\Pi)) = \sum_{w \in \program(\Pi) \ \mid \ \exists m \in AS(w), \ m \models I} P(w),
\end{equation*}
\begin{equation*}
% \label{eq:lower_prob_interpretation}
\lowerprob(I \mid \program(\Pi)) = \sum_{w \in \program(\Pi) \ \mid \ \forall m \in AS(w), \ m \models I} P(w).
\end{equation*}

% The task of parameter learning in a PASP $\program(\Pi)$ aims to find an optimal probability assignment to the values in $\Pi$ such that the product of the lower or upper probability of the interpretation queries is maximized, as stated in Definition~\ref{def:pl_pasp}.

\begin{definition}[Parameter Learning in probabilistic answer set programs]
\label{def:pl_pasp}
Given a PASP $\mathcal{P}(\Pi)$ and a set of (partial) interpretations $\interpretations$, the goal of the parameter learning task is to find a probability assignment to the probabilistic facts such that the product of the lower (or upper) probabilities of the partial interpretations is maximized, i.e., solve:
\begin{equation*}
\Pi^* = \argmax_\Pi \lowerprob(\interpretations \mid \mathcal{P}(\Pi)) =  \argmax_\Pi \prod_{I \in \interpretations} \lowerprob(I \mid \mathcal{P}(\Pi)) 
\end{equation*}
which can be equivalently expressed as
\begin{equation}
\label{eq:learning_task_log}
\Pi^* = \argmax_\Pi \mathrm{log}(\lowerprob(\interpretations \mid \mathcal{P}(\Pi))) = \argmax_\Pi \sum_{I \in \interpretations} \ \mathrm{log}(\lowerprob(I \mid \mathcal{P}(\Pi))) 
\end{equation}
also known as \textit{log-likelihood} (LL).
\revisione{The maximum value of the LL is 0, obtained when all the interpretations have probability 1.
The use of log probabilities is often preferred since summations instead of products are considered, thus possibly preventing numerical issues, especially when many terms are close to 0.}
\end{definition}
Note that, since the probability of a query (interpretation) is described by a range, we need to select whether we maximize the lower or upper probability.
A solution that maximizes one of the two bounds may not be a solution that also maximizes the other bound.
% To see this, suppose that we only have one interpretation $I$ with $\lowerprob(I) = 0$ and $\upperprob(I) > 0$.
\revisione{To see this, consider the program $\{\{q \impl a, b, not\ nq\}, \{nq \impl a, b, not\ q\}\}$ where $a$ and $b$ are both probabilistic with probability $p_a$ and $p_b$, respectively.
Suppose we have the interpretation $I = \langle \{q\}, \{\} \rangle$.
Here, $\lowerprob(I) = 0$ while $\upperprob(I) = p_a \cdot p_b$.
Thus, any probability assignment to $p_a$ and $p_b$ maximizes the lower probability (which is always 0) but only the assignment $p_a = 1$ and $p_b = 1$ maximizes $\upperprob(I)$.
}

\begin{example}
\label{ex:interpretations}
\revisione{Consider the program shown in Example~\ref{ex:running_pasp}. % with an additional probabilistic fact $0.1::edge(2,3)$.
Suppose we have two interpretations: $I_0 = \langle \{path(1,3)\}, \{path(1,4)\} \rangle$ and $I_1 = \langle \{path(1,4)\}, \{\} \rangle$.
Thus, we have two interpretation queries: $q_{I_0} = path(1,3), \ not \ path(1,4)$ and $q_{I_1} = path(1,4)$.
Suppose that the probabilities of all the four probabilistic facts can be set and call this set $\Pi$.
The parameter learning task of Definition~\ref{def:pl_pasp} involves solving: $\Pi^* = \argmax_\Pi (\mathrm{log}( P(q_{I_0} \mid \Pi)) + \mathrm{log} (P(q_{I_1} \mid \Pi)))$.
}
\end{example}

\cite{AzzBelRig2022ParameterLearningPASP} focused on ground probabilistic facts whose probabilities should be learnt and proposed an algorithm based on Expectation Maximization (EM) to solve the task.
Suppose that the target is the upper probability.
The treatment for the lower probability is analogous and only differs in the considered bound.
The EM algorithm alternates an expectation phase and a maximization phase, until a certain criterion is met (usually, the difference between two consecutive iterations is less than a given threshold).
This involves computing, in the expectation phase, for each probabilistic fact $a_i$ whose probability should be learnt:
\begin{equation}
\label{eq:upper_expectation}
\upperexp[a_{i0}] = \sum_{I \in \interpretations} \upperprob(not \ a_{i} \mid I), \ \ \ 
\upperexp[a_{i1}] = \sum_{I \in \interpretations} \upperprob(a_{i} \mid I).
\end{equation}
These values are used in the maximization step to update each parameter $\Pi_i$ as:
\begin{equation}
\label{eq:update_parameters}
\Pi_i = 
  \frac{\upperexp[a_{i1}]}{\upperexp[a_{i0}] + \upperexp[a_{i1}]} = 
  \frac{\sum_{I \in \interpretations} \upperprob(a_{i} \mid I)}{\sum_{I \in \interpretations}\upperprob(not \ a_{i} \mid I) + \upperprob(a_{i} \mid I)}.
\end{equation}

\section{Algorithms for Parameter Learning}
\label{sec:algorithm}
We propose two algorithms for solving the parameter learning task, both based on the extraction of symbolic equations from the \revisione{NNF~\citep{DBLP:journals/jair/DarwicheM02}} representing a query.
So, we first describe this common part.
In the following, when we consider the probability of a query we focus on the upper probability.
The treatment for the lower probability is analogous and only differs in the considered bound.

\subsection{Extracting Equations from a NNF}
\label{subsec:extracting_eq}
The upper probability of a query $q$ is computed as a sum of products (see Equation~\ref{eq:upper_lower_prob}).
If, instead of using the probabilities of the facts, we keep them symbolic (i.e., with their name), we can extract a nonlinear symbolic equation for the query, where the variables are the parameters associated with the learnable facts.
Call this equation $f_{up}(\Pi)$ where $\Pi$ is the set of parameters.
Its general form is $f_{up}(\Pi) = \sum_{w_i}\prod_{a_j \in w_i} p_j \prod_{a_j \not \in w_i} (1-p_j) \cdot k_i$ where $k_i$ is the contribution of the probabilistic facts with fixed probability for world $w_i$.
We can cast the task of extracting an equation for a query as a 2AMC problem.
To do so, we can consider as inner semiring and as transformation function the ones proposed by~\cite{AzzRig2023-AIXIA-IC} and described in Section~\ref{subsec:pasp}.
From this inner semiring we obtain two values, one for the lower and one for the upper probability.
The sensitivity semiring by~\cite{10.1016/j.jal.2016.11.031} allows the extraction of an equation from an AMC task.
We have two values to consider, so we extend that semiring to
$R^o = (\mathbb{R}[X], +, -, (0, 0), (1, 1))$ with
\begin{equation*}
w_o(l) =  
\begin{cases}
(p,p) & \text{for a p.f. } p::a \text{ with fixed probability and } l = a \\
(1-p,1-p) & \text{for a p.f. } p::a \text{ with fixed probability and } l = not \ a \\
(\pi_a,\pi_a) & \text{for a learnable fact } \pi_a::a \text{ and } l = a \\
(1 - \pi_a, 1 - \pi_a) & \text{for a learnable fact } \pi_a::a \text{ and } l = not \ a \\
(1,1) & \text{otherwise}
\end{cases}
\end{equation*}
where p.f. stands for probabilistic fact and $\mathbb{R}[X]$ is the set of real valued functions parameterized by $X$. 
Variable $\pi_a$ indicates the symbolic probability of the learnable fact $a$.
In this way, we obtain a nonlinear equation that represents the probability of a query.
When evaluated by replacing variables with actual numerical values, we obtain the probability of the query when the learnable facts have such values.
Simplifying the obtained equation is also a crucial step since it might significantly reduce the number of operations needed to evaluate it.  

\begin{example}
\label{ex:symbolic_eq}
If we consider Example~\ref{ex:running_pasp} with query $path(1,4)$ and associate a variable $\pi_{xy}$ with each $edge(x,y)$ probabilistic fact (and consider them as learnable), the symbolic equation for its upper probability, \revisione{i.e., $f_{up}$}, is $\pi_{12} \cdot \pi_{24} \cdot \pi_{13} + \pi_{12} \cdot \pi_{24} \cdot (1 - \pi_{13})$, that can be simplified to $\pi_{12} \cdot \pi_{24}$.
\end{example}

We now show how to adopt symbolic equations to solve the parameter learning task.

\subsection{Solving Parameter Learning with Constrained Optimization}
\label{subsec:learning_opt}
% At a high level, the idea behind the first algorithm is the following: the upper probability of a query $q$ is computed as the sum of the probabilities of the worlds that have at least one answer set with $q$ in it (Equation~\ref{eq:upper_lower_prob}).
The learning task requires maximizing the sum of the log-probabilities of the interpretations (Equation~\ref{eq:learning_task_log}) where the tunable parameters are the facts whose probability should be learnt.
\revisione{Algorithm~\ref{alg:learning_opt} sketches the involved steps}.
We can extract the equation for the probability of each interpretation from the NNF and consider each parameter of a learnable fact as a variable (\revisione{function \textsc{GetEquationFromNNF}}).
\revisione{To reduce the size of the equation we simplify it (function \textsc{Simplify}).}
Then, we set up a constrained nonlinear optimization problem where the target is the maximization of the sum of the equations representing the log probabilities of the interpretations (\revisione{function \textsc{SolveOptimizationProblem}}).
We need to also impose that the parameters of the learnable facts are between 0 and 1.
In this way, we can easily adopt off-the-shelf solvers and do not need to write a specialized one.

\begin{algorithm}[t]
\begin{scriptsize}
\caption{\revisione{Function \textsc{LearningOPT}: solving the parameter learning task targeting the upper probability with constrained optimization in a PASP $P$ with learnable probabilistic facts $\Pi$ and  interpretations $I$.}}
\label{alg:learning_opt}
\begin{algorithmic}[1]
\Function{LearningOPT}{$P,I$}
  \State $eqsList \gets \emptyset$
  \For{$i \in I$}
    % \State $\mathit{NNF} \gets$ \Call{BuildNNF}{$i$}
    \State $eq \gets$ \Call{GetEquationFromNNF}{$i$}
    \State $eqsList \gets eqsList \ \cup$ \Call{Simplify}{$eq$}
  \EndFor
  \State $\Pi^* \gets$ \Call{SolveOptimizationProblem}{$eqsList$}
  \State \Return $\Pi^*$
\EndFunction
\end{algorithmic}
\end{scriptsize}
\end{algorithm}

\subsection{Solving Parameter Learning with Expectation Maximization}
\label{subsec:learning_em}
We propose another algorithm that instead performs Expectation Maximization as per Equation~\ref{eq:upper_expectation} and Equation~\ref{eq:update_parameters}.
\revisione{It is sketched in Algorithm~\ref{alg:learning_em}}.
For each interpretation $I$, we add its interpretation query $q_I$ into the program.
To compute $\upperprob{(a_j \mid I_k)}$ for each learnable probabilistic fact $a_j$ and each interpretation $I_k$ we proceed as follows: first, we compute $\upperprob{(a_j, I_k)}$ and $\lowerprob{(not \ a_j, I_k)}$.
Then, Equation~\ref{eq:conditional_probability} allows us to compute $\upperprob(a_j \mid I_k)$.
Similarly for $\lowerprob{(a_j \mid I_k)}$.
We extract an equation for each of these queries (\revisione{function \textsc{GetEquationsFromNNF}}) and iteratively evaluate them until the convergence of the EM algorithm \revisione{(lines~\ref{alg_line:start_em_loop}--\ref{alg_line:end_em_loop} that alternates the expectation phase with function \textsc{Expectation}, the maximization phase with function \textsc{Maximization}, and the computation of the log-likelihood with function \textsc{ComputeLL}).}
\revisione{We consider as default convergence criterion a variation of the log-likelihood less than $5 \cdot 10^{-4}$ between two subsequent iterations.
However, this parameter can be set by the user.}
If we denote with $n_p$ the number of probabilistic facts whose probabilities should be learnt and $n_i$ the number of interpretations, we need to extract equations for $2 \cdot n_p \cdot n_i$ queries.
However, this is possible with only one pass of the NNF: \revisione{across different iterations, the structure of the program is the same, only the probabilities, and thus the result of the queries, will change.
Thus, we can store the performed operations in memory and use only those to reevaluate the probabilities, without having to rebuild the NNF at each iteration}.

\begin{algorithm}[t]
\begin{scriptsize}
\caption{\revisione{Function \textsc{LearningEM}: solving the parameter learning task targeting the upper probability with Expectation Maximization in a PASP $P$ with learnable probabilistic facts $\Pi$ and with interpretations $I$.}}
\label{alg:learning_em}
\begin{algorithmic}[1]
\Function{LearningEM}{$P,I$}
  \State $eqsList \gets$ \Call{GetEquationsFromNNF}{$\pi,I$}
  % \For{$\pi \in \Pi$}
  %   \For{$i \in I$}
  %     \State $v_0 \gets$ \Call{GetEquationFromNNF}{$\pi,i$} \Comment{$\upperprob(\pi, i)$}
  %     \State $v_1 \gets $ \Call{GetEquationFromNNF}{$not \ \pi,i$} \Comment{$\lowerprob(not \ \pi, i)$}
  %     \State $eq \gets \frac{v_0}{v_0 + v_1}$ \Comment{$\upperprob(\pi \mid i)$}
  %     \State $eqsList \gets eqsList \ \cup$ \Call{Simplify}{$eq$}
  %   \EndFor
  % \EndFor
  \State $ll_0 \gets 0$
  \State $ll_1 \gets 1$
  \While {$(ll_1 - ll_0) > 5 \cdot 10^{-4}$} \label{alg_line:start_em_loop} \Comment{Loop until convergence.}
    \State $ll_0 \gets ll_1$
    \State $E \gets$ \Call{Expectation}{$eqsList$}
    \State $\Pi \gets$ \Call{Maximization}{$E$}
    \State $ll_1 \gets$ \Call{ComputeLL}{$eqsList,\Pi$} 
  \EndWhile \label{alg_line:end_em_loop}
  \State \Return $\Pi$
\EndFunction
\end{algorithmic}
\end{scriptsize}
\end{algorithm}

\begin{example}
\revisione{Consider Example~\ref{ex:interpretations} and its two interpretation queries, $q_{I_0}$ and $q_{I_1}$.
If we consider their symbolic equations we have $\pi_{12} \cdot \pi_{24}$ (see Example~\ref{ex:symbolic_eq}) and $\pi_{13} \cdot (\pi_{12} + \pi_{24} - \pi_{12} \cdot \pi_{24})$, where with $\pi_{xy}$ we indicate the probability of $edge(x,y)$.
If we compactly denote with $\Pi$ the set of all the probabilities (i.e., all the $\pi_{xy}$), the optimization problem of Equation~\ref{eq:learning_task_log} requires solving $\Pi^* = \argmax_\Pi (\mathrm{log}( \pi_{12} \cdot \pi_{24}) + \mathrm{log} (\pi_{13} \cdot (\pi_{12} + \pi_{24} - \pi_{12} \cdot \pi_{24})))$.
The optimal solution is to set the values of all the $\pi_{xy}$ to 1, obtaining a log-likelihood of 0.
Note that, in general, it is not always possible to obtain a LL of 0.
}
% 0.99::edge(1,2).
% 0.99::edge(2,4).
% 0.99::edge(1,3).
% path(X,Y):- connected(X,Z), path(Z,Y).
% path(X,Y):- connected(X,Y).

% connected(X,Y):- edge(X,Y).

% q:- path(1,3), not path(1,4).

% query(path(1,4)).
% query(q).

\end{example}

\section{Related Work}
\label{sec:related}
There are many different techniques available to solve the parameter learning task, but most of them only work for programs with a unique model per world: PRISM~\citep{DBLP:conf/iclp/Sato95} was one of the first tools considering inference and parameter learning in PLP.
Its first implementation, which dates back to 1995, offered an algorithm based on EM.
The same approach is also adopted in EMBLEM~\citep{BelRig13-IDA-IJ}, which learns the parameters of Logic Programs with Annotated Disjunctions (LPADs)~\citep{VenVer04-ICLP04-IC}, i.e., logic programs with disjunctive rules where each head atom is associated with a probability, and in ProbLog2~\citep{DBLP:journals/corr/abs-1304-6810} that learn the parameters of ProbLog programs from partial interpretations adopting the LFI-ProbLog algorithm of~\cite{DBLP:conf/pkdd/GutmannTR11}.
LeProbLog~\cite{DBLP:conf/pkdd/GutmannKKR08} still considers ProbLog programs but uses gradient descent to solve the task.

Few tools consider PASP and allow multiple models per world.
dPASP~\citep{geh2023dpasp} is a framework to perform parameter learning in Probabilistic Answer Set Programs but targets the max-ent semantics, where the probability of a query is the sum of the probabilities of the models where the query is true.
They propose different algorithms based on the computation of a fixed point and gradient ascent.
However, they do not target the Credal Semantics.
Parameter Learning under the Credal Semantics is also available in PASTA~\citep{AzzBellRig2022PASTA,AzzBelRig2022ParameterLearningPASP}.
Here, we adopt the same setting (learning from interpretations) but we address the task with an inference algorithm based on 2AMC and extraction of symbolic equations, rather than projected answer set enumeration, which has been empirically proven more effective~\citep{AzzRig2023-AIXIA-IC}.
This is also proved in our experimental evaluation.

Lastly, there are alternative semantics to adopt in the context of Probabilistic Answer Set Programming, namely, P-log~\citep{DBLP:journals/tplp/BaralGR09}, LPMLN~\citep{DBLP:conf/kr/LeeW16}, and smProbLog~\citep{totis_de_raedt_kimmig_2023}.
The relation among these has been partially explored by~\citep{DBLP:conf/aaai/LeeY17} but a complete treatment providing a general comparison is still missing.
The parameter learning task under these semantics has been partially addressed and an in-depth comparison between all the existing approaches can be an interesting future work. 

\section{Experiments}
\label{sec:experiments}
We ran the experiments on a computer with 16 GB of RAM running at 2.40GHz with 8 hours of time limit.
The goal of the experiments is many-fold: 
i) finding which one of the algorithms is faster;
ii) discovering which one of the algorithms better solves the optimization problem represented by Equation~\ref{eq:learning_task_log} (note that the sum of the log probabilities is maximum at 0, and this happens when all the partial interpretations have probability 1);
iii) evaluating whether different initial probability values impact on the execution time; and
iv) comparing our approach against the algorithm based on projected answer set enumeration of~\cite{AzzBelRig2022ParameterLearningPASP} called PASTA. 
\revisione{We considered only PASTA since it is the only algorithm that currently solves the task of parameter learning in PASP under the credal semantics}.
We used aspmc~\citep{DBLP:conf/kr/EiterHK21} as backend solver for the computation of the formula, SciPy version 1.13.0 as optimization solver~\citep{2020scipy}, and SymPy~\citep{meurer2017sympy} version 1.12 to simplify equations.
In this way, the tool is completely open source\footnote{Datasets and source code available at \url{https://github.com/damianoazzolini/aspmc} \revisione{and on Zenodo with DOI \url{10.5281/zenodo.12667046}.}}.
In the experiments, for the algorithm based on constrained optimization (Section~\ref{subsec:learning_opt}) we tested two nonlinear optimization algorithms available in SciPy, namely COBYLA~\citep{Powell1994cobyla} and SLSQP~\citep{Dieter1994slsqp}.
\revisione{COBYLA stands for Constrained Optimization BY Linear Approximation and is a derivative-free nonlinear constrained optimization algorithm based on linear approximation of the objective function and constraints during the solving process.
On the contrary, SLSQP is based on Sequential Least Squares Programming, which is based on solving, at each iteration, a least square problem equivalent to the original problem. }
In the results, we denote the algorithm based on Expectation Maximization described in Section~\ref{subsec:learning_em} with EM.

\subsection{Datasets Descriptions}
\label{subsec:datasets_descriptions}
We considered four datasets.
% Here we report instances for aspmc.
% The ones for PASTA differ only in the negation symbol: \revisione{$\backslash+$ is adopted in aspmc while $not$ in PASTA.}
Where not explicitly specified, all the initial probabilities of the learnable facts are set to 0.5.
For all the instances, we generated configurations with 5, 10, 15, and 20 interpretations.
\revisione{The atoms to be included in the interpretations are taken uniformly at random} from the Herbrand base of each program and some of them are considered as negated, again uniformly at random.

The \textit{coloring} dataset models the graph coloring problem, where each node is associated with a color and configurations are considered valid only if nodes associated with the same color are not connected.
We do not explicitly remove invalid solutions but only mark them as not valid.
\revisione{This makes the following dataset crucial to test our proposal when there are an increasing number of answer sets for each world.}
All the instances have the following rules:
\begin{lstlisting}
red(X) :- node(X), not green(X), not blue(X).
green(X) :- node(X), not red(X), not blue(X).
blue(X) :- node(X), not red(X), not green(X).
e(X,Y) :- edge(X,Y).
e(Y,X) :- edge(Y,X).
c0 :- e(X,Y), red(X), red(Y).
c1 :- e(X,Y), green(X), green(Y).
c2 :- e(X,Y), blue(X), blue(Y).
valid :- not c0, not c1, not c2.
\end{lstlisting}
The goal is to learn the probabilities of the $edge/2$ facts given an increasing number of interpretations containing the observed colors for some edges and whether the configuration is $valid$ or not.
We considered complete graphs of size 4 (\textit{coloring4}) and 5 (\textit{coloring5}).
Here the interpretations have a random length between 3 and 4.

The \textit{path} dataset models a path reachability problem and every instance has the rules described in Example~\ref{ex:running_pasp}.
\revisione{We considered this dataset since it represents a graph structure that can model many problems.}
The goal is to learn the probabilities of the $edge/2$ facts given that some paths are observed.
We considered random connected graphs with 10 (\textit{path10}) and 15 (\textit{path15}) edges.
Here the interpretations have a random length between 1 and 3.

The \textit{shop} dataset models the shopping behaviour of some people, with different products available where some of them cannot be bought together.
An example of instance (with 2 people) is:
\begin{lstlisting}
bought(spaghetti,john):-shops(john), not bought(steak,john).
bought(steak,john):-shops(john), not bought(spaghetti,john).
bought(spaghetti,mary):-shops(mary), not bought(beans,mary).
bought(beans,mary):-shops(mary), not bought(spaghetti,mary).
bought(spaghetti):-  bought(spaghetti,_).
bought(steak):- bought(steak,_).
:- bought(spaghetti), bought(steak).
\end{lstlisting}
The goal is to learn the probabilities of the $shops/1$ learnable facts given an increasing number of products that are observed being bought (or not bought) and an increasing number of people.
We considered instances with 4 (\textit{shop4}), 8 (\textit{shop8}), 10 (\textit{shop10}), and 12 (\textit{shop12}) people.
Here the interpretations have a random length between 1 and 10.
\revisione{With this dataset we assess the learning algorithms in programs when constraints prune some solutions.}

The \textit{smoke} dataset, adapted from~\citep{totis_de_raedt_kimmig_2023}, models a network where some people smoke and others are influenced by the smoking behaviour of their friends.
\revisione{This is a well-known dataset often used to benchmark probabilistic logic programming systems.}
An example of instance with two people is:
\begin{lstlisting}
0.1::asthma_f(1). 0.4::asthma_fact(1). 0.3::stress(1).
0.1::asthma_f(2). 0.4::asthma_fact(2). 0.3::stress(2).
0.2 :: predisposition.

smokes(X) :- stress(X).
smokes(X) :- influences(Y, X), smokes(Y).
asthma_rule(X):- smokes(X), asthma_fact(X).

asthma(X) :- asthma_f(X).
asthma(X) :- asthma_rule(X).

ill(X)  :- smokes(X), asthma(X), not n_ill(X).
n_ill(X):- smokes(X), asthma(X), predisposition, not ill(X).
\end{lstlisting}
Here, the learnable facts have signature $\mathit{influences}/2$ and the goal is to learn their probabilities given that some $ill/1$ facts are observed.
We consider instances with 3 (\textit{smoke2}), 4 (\textit{smoke4}), and 6 (\textit{smoke6}) people.
Here the interpretations have a random length between 1 and 3.

\subsection{Results}
\label{subsec:results}

\begin{table}
\centering
\caption{Final \revisione{log-likelihood} (LL) values for the tested algorithms on six selected instances with the initial probability of the learnable facts set to 0.5.
\revisione{The column \# int. contains the number of interpretations considered, the column EM contains the results obtained with Expectation Maximization, columns C. COBYLA and C. SLSQP stands for constrained optimization solved with, respectively, COBYLA and SLSQP, and the column PASTA contains the results obtained with the PASTA solver.}
}
\label{tab:ll_difference}
\scriptsize
%\begin{tabular}{| c || c | c | c | c | | c | c | c | c |   }
%{\tablefont\begin{tabular}{@{\extracolsep{\fill}}lccccc|cccc}
{\tablefont\begin{tabular}{@{\extracolsep{\fill}}l |  cccc | cccc }
\multicolumn{1}{c}{} & \multicolumn{4}{c}{\textit{coloring4}} & \multicolumn{4}{c}{\textit{shop8}}\\
%\hline
\topline
\# int. \ & EM & C. COBYLA & C. SLSQP & PASTA \ & EM & C. COBYLA & C. SLSQP & PASTA \
\midline
%\hline
5 & -0.016 & 0.000   & 0.000 & -34.539  \  & -1.389  & -12.710 & -1.385 & -34.077 \\
10 & -0.049 & 0.000   & 0.000 & -55.262 \  & -2.257  & 0.000   & 0.000  & -63.268 \\
15 & -0.044 & 0.000   & 0.000 & -96.709 \  & -1.923  & 0.000   & -1.902 & -75.986 \\
20 & -0.030 & -36.775 & 0.000 & -124.340  \ & \revisione{-1.388} & -72.524 & -0.011 & -127.679
%\hline
%\hline
\midline
\multicolumn{1}{c}{} & \multicolumn{4}{c}{\textit{path10}} & \multicolumn{4}{c}{\textit{smoke4}} \\
%\hline
\topline
\# int. \ & EM & C. COBYLA & C. SLSQP & PASTA \ & EM & C. COBYLA & C. SLSQP & PASTA \
\midline
%\hline
5  & -0.002 & 0.000 & 0.000 & -27.631  \ & -9.430 & -9.423  & -9.423  & -20.723  \\
10 & -0.004 & 0.000 & 0.000 & -48.354  \ & -25.164 & -25.160 & -25.160 & -48.354 \\
15 & -0.007 & 0.000 & 0.000 & -75.985  \ & -50.023 & -50.020 & -50.020 & -96.709 \\
20 & -0.004 & 0.000 & 0.000 & -103.616 \  & -64.576 & -64.571 & -64.571 & -110.524\midline
%\hline
%\hline
\multicolumn{1}{c}{} & \multicolumn{4}{c}{\textit{smoke3}} & \multicolumn{4}{c}{\textit{shop4}} \\
%\hline
\topline
\# int. \ & EM & C. COBYLA & C. SLSQP & PASTA \ & EM & C. COBYLA & C. SLSQP & PASTA \
\midline
%\hline
5  & -14.633 & -14.630 & -14.630 & -27.631  \ & -1.462 & -150.606 & 0.000  & -20.633  \\
10 & -36.711 & -36.708 & -36.708 & -69.078  \ & -5.669 & 0.000    & 0.000  & -54.992 \\
15 & -35.172 & -35.166 & -35.166 & -69.078  \ & -7.283 & -49.346  & 0.000  & -74.710 \\
20 & -72.872 & -72.869 & -72.869 & -110.524 \ & -2.770 & 0.000    & -2.768 & -101.824
\midline
\end{tabular}}
\end{table}

\begin{figure}
    \centering
    \begin{subfigure}{0.48\textwidth}
    \centering
    \resizebox{\resizeGraphFactor\textwidth}{!}{
	\includegraphics[scale=.5]{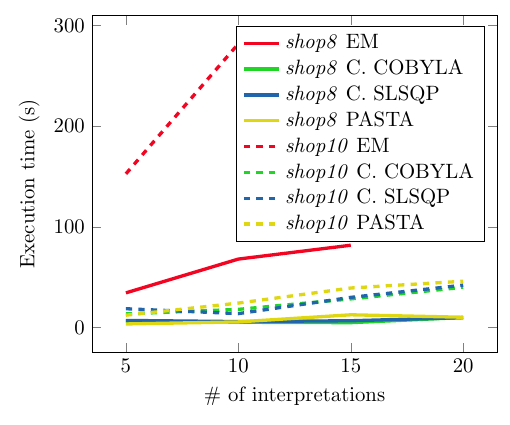}
    }
\caption{\textit{shop8} and \textit{shop10}}
\label{subfig:shop_var_interpretations}
\end{subfigure}%
\hfill
\begin{subfigure}{0.48\textwidth}
\resizebox{\resizeGraphFactor\textwidth}{!}{
\includegraphics[scale=.5]{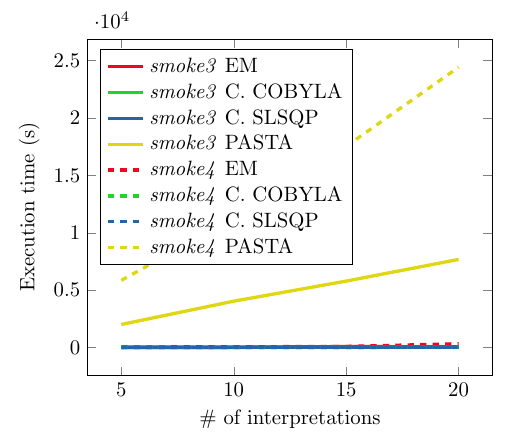}
  }
\caption{\textit{smoke3} and \textit{smokers4}}
\label{subfig:smokers_var_interpretations}
\end{subfigure}

\begin{subfigure}{0.48\textwidth}
    \centering
    \resizebox{\resizeGraphFactor\textwidth}{!}{
   \includegraphics[scale=.5]{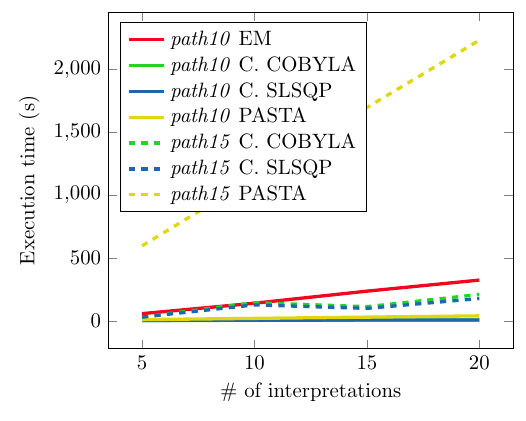}
   }
\caption{\textit{path10} and \textit{path15}}
\label{subfig:path_var_interpretations}
\end{subfigure}%
\hfill
\begin{subfigure}{0.48\textwidth}
\resizebox{\resizeGraphFactor\textwidth}{!}{
   \includegraphics[scale=.5]{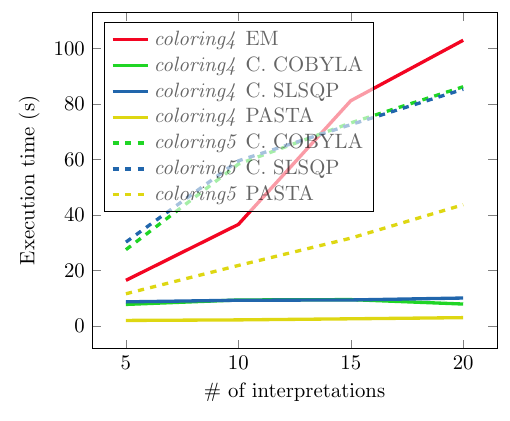}
}
\caption{\textit{coloring4} and \textit{coloring5}}
\label{subfig:coloring_var_interpretations}
\end{subfigure}

\caption{
Execution times for
\textcolor{first_plot}{EM},
\revisione{constrained optimization solved with}
\textcolor{second_plot}{COBYLA} and
\textcolor{third_plot}{SLSQP}, and
\textcolor{fourth_plot}{PASTA},
by increasing the number of interpretations.
For \textit{path15} and \textit{coloring5} the line for EM is missing since it cannot solve any instance.
The initial probabilities for learnable facts are set to 0.5.
}
\label{fig:exec_time_algorithms}
\end{figure}

\begin{figure}
\centering
\begin{subfigure}{0.48\textwidth}
\centering
\resizebox{\resizeGraphFactor\textwidth}{!}{
     \includegraphics[scale=.5]{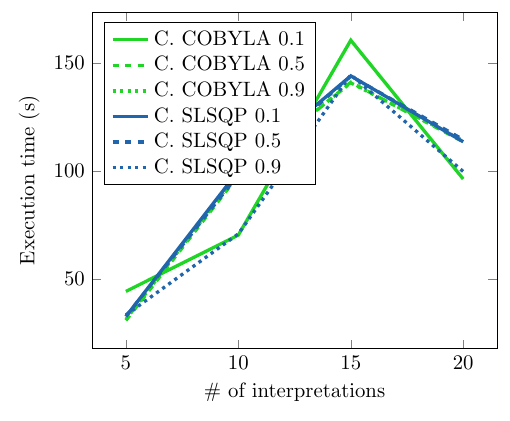}
}
\caption{\textit{shop12}}
\label{subfig:shop12_init_prob}
\end{subfigure}%
\hfill
\begin{subfigure}{0.48\textwidth}
\resizebox{\resizeGraphFactor\textwidth}{!}{
\includegraphics[scale=.5]{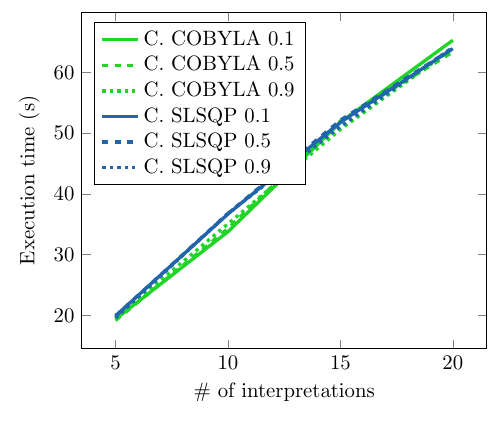}
}
\caption{\textit{smoke6}}
\label{subfig:smoke6_init_prob}
\end{subfigure}

\begin{subfigure}{0.48\textwidth}
    \centering
    \resizebox{\resizeGraphFactor\textwidth}{!}{
    \includegraphics[scale=.5]{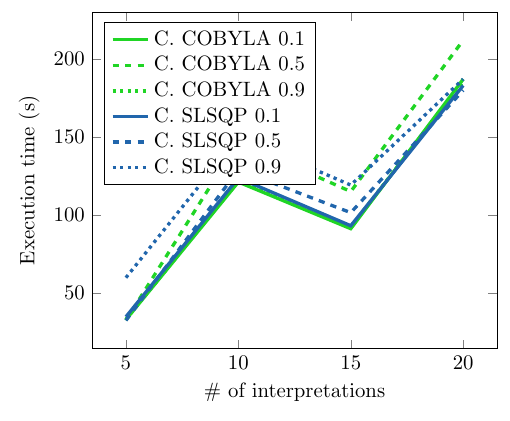}
    }
\caption{\textit{paths15}}
\label{subfig:paths15_init_prob}
\end{subfigure}%
\hfill
\begin{subfigure}{0.48\textwidth}
\resizebox{\resizeGraphFactor\textwidth}{!}{
    \includegraphics[scale=.5]{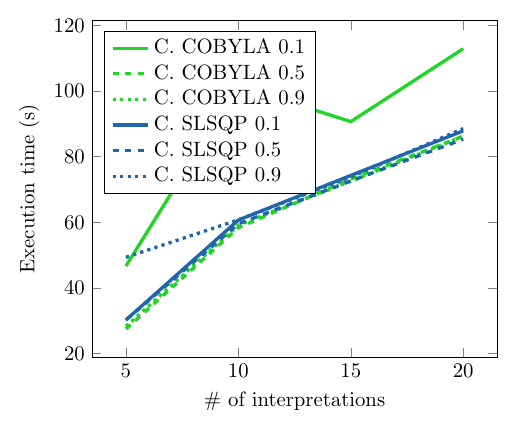}
    }
\caption{\textit{coloring5}}
\label{subfig:coloring5_init_prob}
\end{subfigure}

\caption{
Execution times of \revisione{constrained optimization solved with} \textcolor{second_plot}{COBYLA} and \textcolor{third_plot}{SLSQP} on \textit{coloring5}, \textit{path15}, \textit{shop12}, and \textit{smoke6} with 0.1, 0.5, and 0.9 as initial values for the learnable facts.
}
\label{fig:plot_compare_probs}
\end{figure}

First, we compare the execution times of the four proposed algorithms.
Results are reported in Figure~\ref{fig:exec_time_algorithms}.
COBYLA and SLSQP have comparable execution times while EM and PASTA are often the slowest.
The only instances where PASTA is faster are \textit{coloring4} and \textit{coloring5}.
Given the imposed time and memory limits, EM cannot solve the instances \textit{shop12} with any number of interpretations and \textit{shop10} with 15 and 20 interpretations, and \textit{shop8} with 20 interpretations, \textit{coloring5} with any number of interpretations, and \textit{path15} with any number of interpretations (all due to memory limit) while PASTA was not able to solve \textit{smoke6} with any number of interpretations and \textit{smoke5} with 10, 15, and 20 interpretations (all due to time limit).
The two algorithms based on constrained optimization are able to solve every considered instance.

We also evaluated the algorithms in terms of final log-likelihood.
Table~\ref{tab:ll_difference} shows the results on the \textit{coloring4}, \textit{path10}, \textit{shop4}, \textit{shop8}, \textit{smoke3}, and \textit{smoke4} instances.
SLSQP has the best performance: it seems to be able to better maximize the LL in \textit{coloring4}, \textit{shop4}, and \textit{shop8}, while COBYLA cannot reach such maxima.
For the other two, the results are equal.
EM is also competitive with SLSQP.
PASTA has the worst performance for all the datasets \revisione{and cannot reach a LL of 0 while other algorithms, such as the ones based on constrained optimization, succeed.}

Lastly, we also run experiments with the two optimization algorithms by considering different initial probability values.
Figure~\ref{fig:plot_compare_probs} shows the results.
In general, the initial probability values have little influence on the execution time for SLSQP.
The impact is more evident on COBYLA: for example, for \textit{coloring5} with 10 interpretations, setting the initial values to 0.1 requires 100 seconds of computation while setting them to 0.5 or 0.9 requires 60 seconds.
We extended this evaluation also to EM, but we do not report the results since the initial value makes no difference in terms of execution time.
In the current version, PASTA does not allow setting an initial probability value different from 0.5.

Overall, the algorithm based on constrained optimization outperforms in terms of efficiency and in terms of final log-likelihood both the algorithm based on EM and PASTA.
EM also outperforms PASTA in terms of final LL but it often requires excessive memory and cannot solve instances solvable by PASTA.
\revisione{Nevertheless, several improvements may be integrated within our approach to possibly increase the scalability: 
i) considering different alternatives for the representation of symbolic equations, possibly more compact, also considering simmetries~\citep{AzzRig23-IJAR-IJ};
ii) developing ad-hoc simplification algorithms to reduce the size of the symbolic equations since they are only composed by summations of products.}
\revisione{Furthermore, considering higher runtimes could be beneficial for enumeration-based techniques. }

% \commento{
%   confronto tempi di esecuzione con valore iniziale 0.5 tra tutti gli algoritmi - ora figura 2 che diventa figura 1
% }

% \commento{
%   confronto LL tra tutti gli algoritmi: prob iniziale 0.5. tabella con dataset e algoritmi sulle righe e LL sulle colonne. vedere dove mettere interpretazioni. 14 configurazioni - vedere come viene
% }

% \commento{per tempo di esecuzione con diverse probabilità iniziali (cobyla e slsqp): 
% shop 12 (0.1, 0.5, 0.9 e 1 se ci sta), 
% somkers 6 (0.1, 0.5, 0.9 e 1 se ci sta) 
% paths 15
% coloring 5 - ora figura 1}

% \commento{
%   come è suddiviso il tempo di esecuzione nelle varie parti dell'algoritmo: in sospeso per ora.
% }

\section{Conclusions}
\label{sec:conclusions}
In this paper, we propose two algorithms to solve the task of parameter learning in PASP.
Both are based on the extraction of a symbolic equation from a compact representation of the problem but differ in the solving approach: one is based on Expectation Maximization while the other is based on constrained optimization.
For the former, we tested two algorithms, namely COBYLA and SLSQP.
We compare them against PASTA, a solver adopting projected answer set enumeration.
Empirical results show that the algorithms based on constrained optimization are more accurate and also faster than the ones based on EM and PASTA.
Furthermore, the one based on EM is still competitive with PASTA but often requires an excessive amount of memory.
As a future work we plan to consider the parameter learning task for other semantics \revisione{and also theoretically study the complexity of the task in light of existing complexity results for inference~\citep{cozman2020complexity}}.

\subsection*{Competing interests}
The authors declare none.

\section*{Acknowledgements} 
This work has been partially supported by Spoke 1 ``FutureHPC \& BigData'' of the Italian Research Center on High-Performance Computing, Big Data and Quantum Computing (ICSC) funded by MUR Missione 4 - Next Generation EU (NGEU) and by Partenariato Esteso PE00000013 - ``FAIR - Future Artificial Intelligence Research'' - Spoke 8 ``Pervasive AI'', funded by MUR through PNRR - M4C2 - Investimento 1.3 (Decreto Direttoriale MUR n. 341 of 15th March 2022) under the Next Generation EU (NGEU).
All the authors are members of the Gruppo Nazionale Calcolo Scientifico -- Istituto Nazionale di Alta Matematica (GNCS-INdAM).
% We acknowledge the CINECA award under the ISCRA initiative, for the availability of high-performance computing resources and support.
Elisabetta Gentili contributed to this paper while attending the PhD programme in Engineering Science at the University of Ferrara, Cycle XXXVIII, with the support of a scholarship financed by the Ministerial Decree no. 351 of 9th April 2022, based on the NRRP - funded by the European Union - NextGenerationEU - Mission 4 ``Education and Research'', Component 1 ``Enhancement of the offer of educational services: from nurseries to universities'' - Investment 4.1 ``Extension of the number of research doctorates and innovative doctorates for public administration and cultural heritage".

\bibliographystyle{apalike}
% \bibliography{
%   bibtexrepository/journals_long,
%   bibtexrepository/booktitles_long,
%   bibtexrepository/booktitles_springer,
%   bibtexrepository/series_long,
%   bibtexrepository/series_springer,
%   bibtexrepository/publishers_long,
%   bibtexrepository/bibl
% }
\bibliography{biblio_onefile}

\end{document}